# Real-time Secondary Crash Likelihood Prediction Excluding Post Primary Crash Features

Lei Han*, Mohamed Abdel-Aty, *Senior Member, IEEE*, Zubayer Islam, *Member, IEEE*, and Chenzhu Wang

*Abstract*—Secondary crash likelihood prediction is a crucial component of active traffic management system to help mitigate congestion and adverse effects caused by secondary crashes. However, existing methods mainly relied on several post-crash features (e.g., crash type and severity) that are rarely available in real-time, making them impractical for real-world applications. To address such issues, we propose a secondary crash likelihood prediction hybrid model excluding any post-crash features. Specifically, a dynamic spatial-temporal window is designed to extract real-time traffic flow and environmental features among primary and secondary crash segments. Among the hybrid model system, a primary crash prediction model is trained to predict the likelihood of crash leading to secondary crashes. Two models (1 and 2) are proposed to predict the secondary crash likelihood at primary crash and its upstream segments. Model 1 assesses traffic status before secondary crashes against those at normal crashes without secondary crashes, while Model 2 compares the traffic status before secondary crashes with crash-free conditions. For the three models, an ensemble method of six machine learning models is developed to improve prediction accuracy. A voting-based strategy is utilized to combine three models' results to predict the final secondary crash likelihood. Experiments with Florida freeways show that the proposed hybrid model correctly identify 91% of unsafety crashes with low false alarm rate of 0.20 without post-crash features. Model Area Under ROC Curve (AUC) improves from 0.654, 0.744, and 0.902 of the three separate models to 0.952 with the hybrid model—surpassing the performance reported in previous studies.

## I. INTRODUCTION

SECONDARY crashes are defined as those crashes that occur within the spatial and temporal ranges primary crashes. It is estimated that on freeway systems, about 1~2% of the total crashes would lead to secondary crashes [1, 2]. The occurrence of secondary crashes not only reduces the freeway capacity and increases traffic congestion but also causes greater crash severity and loss to people and property. According to the Federal Highway Administration (FHWA), secondary crashes account for 25% of traffic delays and cause approximately 18% of deaths among all freeway crashes in the US [3]. Given the significant economic and social costs, secondary-crash-related indexes (e.g., likelihood, frequency) have been widely used to evaluate the safety level of freeway networks by FHWA and other transportation agencies around the world [3, 4]. Therefore, conducting secondary crash mitigation to improve traffic safety has become one of the priorities for Traffic Incident Management [5, 6].

To prevent the occurrence of secondary crash, considerable effort has been dedicated to identifying the contributing factors of secondary crashes in the last decades [4, 5, 7-10]. For instance, weather, traffic congestion, and several primary crash related features (e.g., crash type, duration, and severity) have been identified as critical contributing factors [4]. Nevertheless, the limitation of such studies is that they are essentially passive post-event analyses, seeking solutions only after secondary crash have occurred [1, 2, 6]. To overcome such disadvantages, that is, shift from passive responses to active prevention, real-time secondary crash likelihood prediction has been proposed as a crucial solution. The underlying concept is that secondary crashes are more likely to occur under certain unsafe traffic situations (e.g., high volume, turbulent traffic, and adverse weather) compared to other crash-free and primary-crash-only cases [1, 2, 11, 12]. By capturing these "high-risk" traffic patterns/precursors, it is possible to predict the likelihood of secondary crashes in a short period (e.g., $5 \sim 10$ min) using the real-time traffic dynamics [2, 11, 12]. Unlike traditional methods, secondary crash prediction model can proactively identify potential time and locations for secondary crashes before they occur. Therefore, they can be implemented in a proactive traffic safety management system to trigger active traffic management strategies, reduce the response time to crashes, and prevent the occurrence of secondary crashes.

Although several secondary crash prediction methods have been developed in existing studies, there are still several issues remaining to be solved. First, existing methods rely on post-crash features such as primary crash type, duration, and severity (Table I). However, these features are rarely available in real-time, making these methods impractical for realistic applications. For instance, the duration and severity of primary crash have been commonly used during modeling procedure. These features are not available in real-time as they require verification from police and transport agencies, which usually takes several days after the crash [13, 14]. To make the model implementable into a real-time system, it is essential to explore new approaches to predict secondary crash likelihood excluding these unavailable post-crash features. Secondly,

This work was supported by the FDOT ATTAIN program. *(Corresponding author: Lei Han).*

Lei Han, Mohamed Abdel-Aty, Zubayer Islam, and Chenzhu Wang are with the Smart and Safe Transportation lab at the Department of Civil, Environmental and Construction Engineering, University of Central Florida, USA (e-mail: le966091@ucf.edu; m.aty@ucf.edu; zubayer.islam@ucf.edu; chenzhu.wang@ucf.edu).

secondary crashes are mainly caused by fluctuating traffic and congestion between the primary and secondary crash segments [4, 15]. Most existing methods focus only on the traffic flow features at the primary crash segment, neglecting those at the upstream secondary crash segments [2]. Such lack of upstream traffic flow features results in poor prediction performance. How to effectively extract traffic status for both primary and secondary crash segments still need to be investigated.

TABLE I
VARIABLES IN EXISTING SECONDARY CRASH STUDIES

| Study | Variables |
|---|---|
| [16] | **(1) Primary Crash features**: crash type, severity, duration, involved vehicle number and type<br>**(2) Spatial features**: peak hour<br>**(3) Traffic flow features**: unsafe speed, traffic shockwave<br>**(4) Environmental features**: weather, lighting, road surface |
| [16, 17] | **(1) Primary Crash features**: crash severity, involved vehicle type<br>**(2) Spatial features**: peak hour<br>**(3) Traffic flow features**: AADT<br>**(4) Road geometric features**: speed limit, type of median, segment length, right shoulder width |
| [11, 18] | **(1) Primary Crash features**: location, crash type, involved vehicle number, type, clearance time<br>**(2) Spatial features**: time of day<br>**(3) Traffic flow features**: traffic condition |
| [1] | **(1) Primary Crash features**: crash type, severity<br>**(2) Spatial features**: peak hour<br>**(3) Traffic flow features**: 5-min average/std/diff between adjacent lanes of speed/occupancy/volume<br>**(4) Environmental features**: weather, lighting, road surface<br>**(5) Road geometric features**: number of lanes, width, curve |
| [2] | **(1) Primary Crash features**: crash severity<br>**(2) Traffic flow features**: 5-min average/std of speed/occupancy/volume |

With the above-mentioned gaps, this study aims to develop a real-time secondary crash likelihood prediction model without relying on any post primary crash information, therefore enabling its integration into real-time proactive traffic management system. Main contributions are as follows:

1. Instead of using the unavailable post-crash features, real-time traffic flow, weather and road geometrics features in dynamic spatial-temporal windows among primary and secondary crash segments were extracted.

2. Improving prediction accuracy through a hybrid model system: In addition to a primary crash prediction model, two distinct models were proposed to predict the likelihood of secondary crash at upstream segments. Model 1 assesses the traffic flow status before secondary crashes against those at crashes without secondary crashes, while Model 2 compares the traffic flow status before secondary crashes with crash-free conditions.

3. To overcome the shortcomings of unstable predictions of a single model, an ensemble method of multiple machine learning models (e.g., XGBoost, Random Forest, and CNN models) was used to improve the accuracy and stability of the proposed model.

The remaining sections of this paper are organized as follows. Section II provides a review of relevant literatures. Section III describes the data preparation. Section IV shows the details of the proposed methodology and section V illustrates the experiment results. Section VI includes discussions about model details. Finally, Section VII offers conclusions and suggestions for future research and practices.

II. LITERATURE REVIEW

A. Secondary Crash Identification

There were generally two methods to determine the impact scope of primary crashes and identify the corresponding secondary crashes:

**Static methods** use fixed temporal and spatial thresholds to depict the impact range of primary crashes. For example, Raub [19] first tried to use 1 mile upstream of a crash and 15 min after the crash as the spatial-temporal thresholds; Moore et al. [20] suggested thresholds of 2 miles and 2 hours after crashes for Los Angeles freeways. Although static methods are easy to implement, the lack of unified standards for these thresholds can significantly affect identification accuracy [12]. Additionally, static methods fail to consider the impact of different crash scenarios. As the primary crash impact areas may vary significantly depending on many factors (e.g., weather, traffic conditions, and time of the day), using static spatial-temporal thresholds results in a high likelihood of misidentification [4].

**Dynamic methods** have been widely utilized to better capture the impact areas of primary crashes and identify secondary crashes in recent studies [8, 9, 18, 21-23]. In these methods, dynamic impact areas of primary crashes are estimated through different approaches, which can be mainly divided into three categories: 1) The queue estimation approach aims to estimate the queue caused by primary crashes, with crashes occurring within the queue identified as secondary crashes [8, 22]. While providing an accurate impact range, it is challenging to obtain high-quality queue data, which requires a large number of cameras or other detectors; 2) The shock wave approach assumes that the impact area forms a spatial-temporal triangular shape by the backward and discharging shockwaves associated with the occurrence and clearance of the primary crash [15-17, 23]. However, this method relies on the key assumption that the transmission of traffic wave is constant and stable, which may not be suitable for the complex and changing conditions of actual traffic flow [4]; 3) The speed contour plot approach has become more popular in recent studies [1, 2, 6, 24]. Empowered by various sensor technologies, it establishes a speed contour map based on the speed measurements of high resolution spatial-temporal grids and estimate the impact ranges by comparing the speed before and after the crashes. Crashes occurring in abnormal areas after a crash can be identified as secondary crashes. Furthermore, the speed on crash-free days were also considered to adjust the speed contour plot to exclude the impact of recurrent congestions [4]. Given the high identification accuracy and flexibility in avoiding specific assumption, this method has been widely used in recent studies for secondary crashes identification [2, 6, 12].

### B. Secondary Crash Prediction

By establishing the relationships between the occurrence of secondary crash and their influencing factors, a variety of secondary crash prediction studies have been conducted in recent decades. In the early stages, researchers focused on exploring the impact of different contributing factors to the likelihood of secondary crashes [7, 9, 15, 24]. Most of primary crash related features have been identified to significantly affect the likelihood of the secondary crashes. For example, Karlaftis et al. [7] found that crashes occurring on weekdays are more likely to lead to secondary crashes than those on weekends. Khattak et al. [25] suggested that the number of involved vehicles and duration of primary crashes are significantly positive with the likelihood of secondary crashes. Moreover, factors related with traffic flow (e.g., Annual Average Daily Traffic (AADT), travel speed), weather (e.g., rainy day, snow), lighting conditions, and road geometric features (speed limit, segment length) were also found to be highly correlated with the secondary crash occurrence [1, 15, 24-26].

Recently, real-time secondary crash likelihood prediction has been proposed to predict the likelihood of secondary crashes in a short period (e.g., 5 ~ 10 min) based on real-time traffic data [2]. For example, Xu et al. [1] extracted the loop detector traffic flow data, geometric characteristics, and primary crash features and developed a logistic regression (LR) model with random effect to predict real-time secondary crash likelihood, which could predict 66% of secondary crashes with the false alarm rate at 0.2; Park and Haghani [12] developed a real-time secondary crash likelihood model based on traffic conditions and the duration of the primary crashes. Two machine learning methods (i.e., Bayesian neural networks (BNN) and Gradient Boosted Decision Trees (GBDT)) were used for prediction and the overall accuracy can reach 82.1%; Li and Abdel-Aty [2] proposed a hybrid real-time secondary crash prediction framework by combining a primary crash and a secondary crash prediction model, which significantly improved the prediction performance (AUC = 0.89) than single models. A recent study [27] proposed a transformer-based secondary crash prediction model and can correctly classify 76.6% secondary crashes.

In summary, the studies on secondary crash contributing factors analysis can reveal specific potential factors and scenarios that can lead to secondary crashes, therefore providing targeted suggestions for secondary crash prevention.

Although their results provided useful insights into the secondary crash mechanism, such post-crash analysis models cannot be applied in the real-time and proactive traffic management system. To meet the needs of real-time secondary crash prevention, a few studies started to use the real-time traffic conditions to predict the real-time secondary crash likelihood and achieve satisfactory predictive performance. However, there are still several research limitations: First, existing models were heavily dependent on post primary crash features, such as crash type, duration, and severity, which were rarely available in real-time [2, 4]. For instance, the duration and severity of primary crashes were widely used in existing studies [1, 2, 10, 16, 25]. However, such features usually require verification from traffic management departments after the crash processing, which are not available for the application in a real-time system. Second, the secondary crash likelihood is highly influenced by the traffic conditions at secondary crash segment, which were ignored by most of studies. Li and Abdel-Aty, [2] used the traffic flow data at secondary crash segment. However, they still rely on the primary crash features (crash severity), which makes the model inappropriate for real-time application.

## III. DATA PREPARATION

### A. Data Collection

In this study, three freeways (I-4, I-75, and I-95) in Central Florida, US were chosen as the study area, with is a total of near 481 miles long in both directions as shown in Fig. 1. According to the road geometry and detector locations, the studied freeways were divided into 1278 segments with the average length of around 0.39 miles. Four kinds of data were collected in this study:

*1) Crash Data:* A total of 21,236 crashes for 4 years (2019-2022) were collected from the Signal Four Analytics (SA4) system, which contains comprehensive details (e.g., crash time, location, type, severity, etc.) on reported crashes across the state of Florida.

*2) Traffic Data:* It was collected from the microwave vehicle detection systems (MVDSs) installed on the freeways. It includes the traffic flow features such as traffic speed, volume, and occupancy of each lane every 30s. The raw traffic data were aggregated over 5 min.

*3) Weather data:* Data was collected from the nearest meteorological stations including the values of temperature, relative humidity, heat index, wind speed, visibility, etc.

*4) Road geometry data:* For each road segment, its geometry data were created to record their geometry information by ArcGis. Multiple features such as segment ID, latitude and longitude of starting and end points, road type, speed limit and length were included.

### B. Secondary Crash Identification

In this study, the speed contour plot method was used to identify the secondary crashes. The basic idea is to identify the spatio-temporal ranges of each crash by comparing the speed of primary crash segment and its upstream segments with that of the same segments during crash-free days. Crashes that occurred within the ranges would be paired as corresponding secondary crashes. The detailed steps are as follows:

Step 1: Create the crash-free speed dataset. To reflect the traffic speed of each segment during normal traffic conditions, the mean and standard deviation of traffic speed at each crash segment were estimated during the crash-free days.

Step 2: Determine thresholds of spatial-temporal range. In previous secondary crash studies, spatial thresholds have commonly been set at 1–2 miles upstream, and temporal thresholds at 0.25–2 h after the primary crash [2, 6-8, 24]. To avoid missing identifications and improve the efficiency of identification, the maximum values in previous studies, with a distance gap of 2 miles and a time gap of 2 h, were chosen as the static spatial-temporal thresholds for initial filtering.

Step 3: Identify the secondary crashes in impact ranges. It is important to exclude the impact caused by recurrent congestion when estimating the impact ranges of a primary crash [1, 10]. Therefore, the differences ($\Delta s_j(t_i)$) between the speed before/after the crash and the speed of crash-free days were calculated for each road segment using Equation (1).

$$\Delta s_j(t_i) = s_j(t_i) - (\overline{s_j}(t_i) - \alpha * \sigma_{s_j}(t_i)) \qquad (1)$$

where $s_j(t_i)$ is the speed of road segment $j$ at time $t_i$, $\overline{s_j}(t_i)$ and $\sigma_{s_j}(t_i)$ denotes the mean and standard deviation of speed for the same road at the same time. Referring to previous studies [2, 27] and our preliminary testing, $\alpha$ is set as 0.25 to reflect normal traffic speed fluctuation range and clearly seperate congestions. Once the speed difference $\Delta s_j(t_i)$ is smaller than 0, it means that segment $j$ is under the impact of crash at time $t_i$. Crashes that happened at such impact spatio-temporal area would be identified as secondary crashes.

Fig. 2 illustrates an example of secondary crash identification using the proposed speed contour plot method. Fig. 2(a) and Fig. 2(b) shows the spatial-temporal distribution of original speed and speed difference, respectively. The first crash happened at 7:20 at segment 138. It then affected some of its upstream segments from 7:20 to 8:35. The speed after the crash dropped significantly than the normal speed (Fig. 2(b)). Those speed contour plots show the propagation of the impact of a crash in both spatial and temporal dimensions. Then at 7:50, another crash happened in its upstream segment 135 within the impact area of a prior crash. Therefore, it was considered as a secondary crash (blue pentagram) to the primary crash (blue circle).

Finally, 513 secondary crashes and 486 primary crashes were identified using the speed contour plot method. The rest 20,237 crashes were defined as "normal crashes" (i.e., crashes without secondary crashes). To the best of our knowledge, the number of secondary crashes in this study is the largest among existing secondary crash studies (e.g., 490 in [6]; 310 in [2]; 204 in [15]).The ratio of secondary crashes to the total number of crashes is 2.46%, which is consistent with the findings of existing studies of 1.98% [6], 1.2% [1], and 1.6% [2].

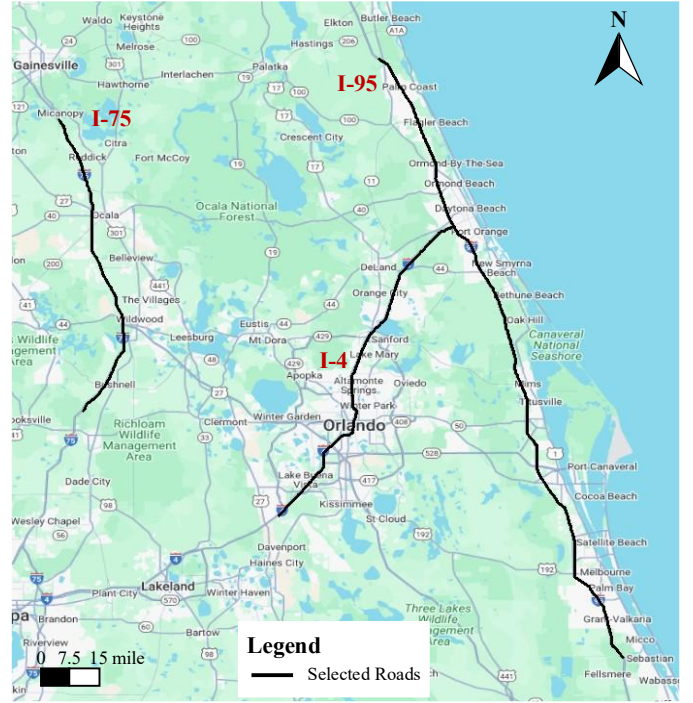

**Fig. 1.** Study area of Florida freeways.

(a) Original speed distribution

(b) Speed difference distribution

**Fig. 2.** Secondary crash identification by speed contour plots.

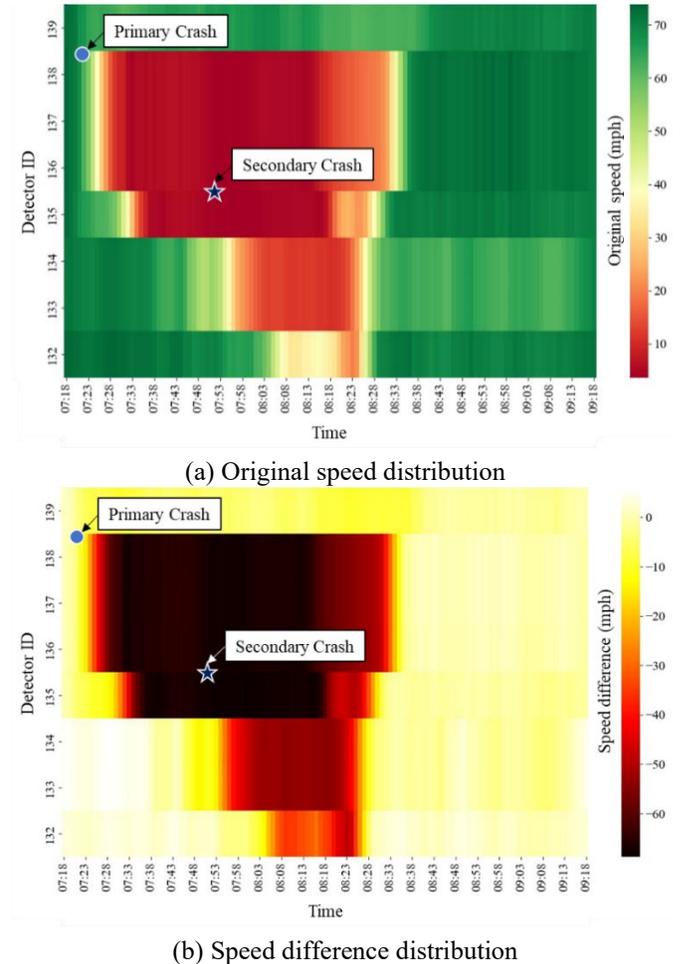

*C. Modeling Variable Extraction*

After obtaining the secondary crashes, real-time variables were extracted as model inputs to predict the secondary crash likelihood. To comprehensively capture the contributing factors to the occurrence of secondary crashes, multiple types of variables were extracted including the traffic flow conditions, weather features, and road geometrics characteristics as shown in Table II. These variables are aggregated into 5-min time slices and matched into each road segment at the spatial dimension. Therefore, there are a total of 22 variables in each temporal-spatial window.

Compared to the traffic flow and crash-related features in previous studies [1, 2, 6, 15], these variables are all available in real time and reflect rich features from static roads to dynamic traffic flow and environment, thus enable the real-time implementation of the proposed method in a proactive traffic management system.

## IV. METHODOLOGY

Fig. 3 is the overall framework of the proposed secondary crash prediction method, extending our previous work [2]. Typically, crash scenarios can be classified into two cases: In case 1, a crash (i.e., non-primary crash) that occurred on segment 1 does not cause a secondary crash on segment 2. While in case 2, a crash (i.e., primary crash) on segment 1 caused a secondary crash on segment 2. The objective of our method is to accurately identify case 2 and determine the secondary crash locations. To achieve this, once a crash occurs at segment 1, a primary crash prediction model is triggered to assess the likelihood that the crash will lead to secondary crashes, i.e., classifying it as a primary or non-primary crash. However, it cannot identify which upstream segments are likely to experience secondary crashes, and its accuracy is limited by the lack of available crash information (e.g., crash severity, clearance time). Thus, secondary crash prediction models are simultaneously implemented at the upstream segments (e.g., segment 2). Taking upstream segment data as input, they can evaluate the secondary crash likelihood for each segment, thereby pinpointing potential secondary crash locations. Finally, a hybrid model would combine these models' results to enhance prediction accuracy and generate the final secondary crash likelihood for these segments.

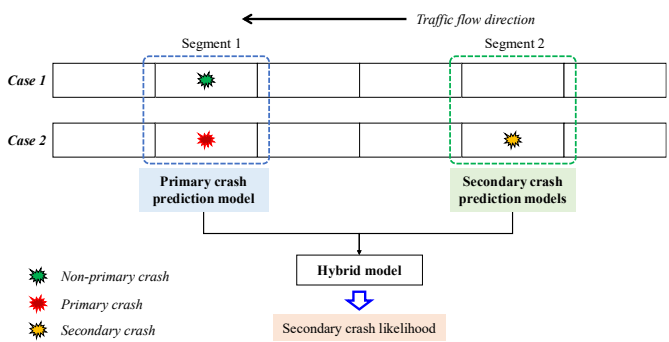

**Fig. 3.** Overall framework of the secondary crash prediction.

| Category | Variable | Description |
|---|---|---|
| **Traffic flow conditions** | Avg_Speed | Average vehicle speed during 5-min period (mile/h) |
| | Avg_Occupancy | Average vehicle occupancy during 5-min period (%) |
| | Avg_Volume | Average vehicle volume during 5-min period (veh/30s) |
| | Std_Speed | Standard deviation of vehicle speed during 5-min period (mile/h) |
| | Std_Occupancy | Standard deviation of vehicle occupancy during 5-min period (%) |
| | Std_Volume | Standard deviation of vehicle volume during 5-min period (veh/30s) |
| | Cv_Speed | Coefficient of variation of vehicle speed during 5-min period (-) |
| | Cv_Occupancy | Coefficient of variation of vehicle occupancy during 5-min period (-) |
| | Cv_Volume | Coefficient of variation of vehicle volume during 5-min period (-) |
| **Weather features** | Temperature | Environment temperature in degrees Fahrenheit (°F) |
| | Humidity | Measurement of environment humidity (%) |
| | Wind_Speed | Wind speed at road segment (m/s) |
| | Precipitation | Measurement of rainfall at road segment (mm) |
| | Visibility | Maximum distance that the human can see (mile) |
| | Condition_Clear | 1 = Clear; 0 = otherwise |
| | Condition_Rain | 1 = Rain; 0 = otherwise |
| **Geometric characteristics** | Speed_Limit | Speed limit value of road segment: 60, 65, 70 (mile/h) |
| | Lane_Count | Count number of lanes of road segment: 2-6 |
| | Segment_Basic | 1 = Basic; 0 = otherwise |
| | Segment_Weaving | 1 = Weaving; 0 = otherwise |
| | Segment_Merge | 1 = Merge; 0 = otherwise |
| | Segment_Diverge | 1 = Diverge; 0 = otherwise |
| | Miles | The length of road segment (mile) |

*: 22 variables are extracted for each road segment.

## A. Primary Crash Prediction Model (PC model)

Fig. 4 shows the development of primary crash prediction model. The basic idea is that traffic conditions, weather, and road geometry may differ between non-primary and primary crash cases. PC model is designed to capture such differences to classify primary crashes from all crash samples. To this end, all primary crash cases (e.g., case 2 in Fig. 4) were labeled as positive samples (Y=1), while non-primary crash cases (e.g., case 1 in Fig. 4) were labeled as negative samples (Y=0). Pre-crash 5-10min variables from the crash and its adjacent segment detectors were utilized as model inputs [1, 2, 11]. Through comparing these positive and negative samples, PC model can learn the feature distributions of primary crashes. As for the modeling algorithms, machine learning (ML) models were chosen due to their high prediction performances, with detailed information provided in the following section.

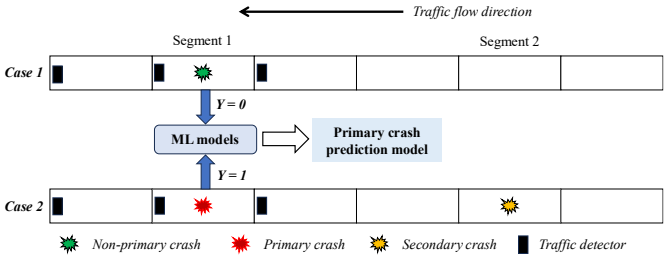

**Fig. 4.** Illustration of primary crash prediction model.

## B. Secondary Crash Prediction Models (SC models)

Existing studies have indicated that secondary crashes are more likely to occur under the congested and disordered traffic caused by prior crashes [4, 11, 27]. Based on this assumption, the key concept of the secondary crash prediction model (SC model) is to capture the differences between the turbulent traffic statues prior secondary crash and normal conditions. Unlike the PC model, SC models are designed for deployment at both crash and upstream segments, with real-time updated after a crash. Therefore, they can dynamically monitor secondary crash risk along the traffic dynamics in upstream segments following the primary crashes.

To develop such SC model, a novel dynamic spatial-temporal windows of traffic data extraction were proposed to effectively describe the turbulent traffic flow states among secondary and primary crash segments. Considering different case-control scenarios for secondary crashes, two distinct SC models were developed to predict secondary crash likelihoods compared to normal crash and non-crash statuses, respectively.

### 1) Dynamic spatial-temporal window of traffic extraction

Taking a typical secondary crash scenario as an example, the occurrence of primary crash may reduce the traffic capacity and generate congestion shockwave propagating to its upstream. Affected by the shockwave, traffic flows on upstream segments become disordered from normal status, eventually leading to secondary crashes. Existing studies mainly adopt a fixed window (red area in Fig. 5(a)) that only focuses on the secondary crash locations, which can hardly capture such traffic flow disorders.

To capture the traffic flow dynamics from the primary crash segments to the secondary crash segments, a dynamic spatial-temporal window is proposed to extract corresponding traffic (blue area in Fig. 5(a)). The time range of 5-25 min before the secondary crash is selected. For example, a secondary crash occurred at 7:53 a.m. in the real case in Fig. 5(b), then the time window is set at 7:28-7:48 a.m. For each time window, the traffic flow data of the center segment and its up and downstream segments would be extracted. As the time window moves from 1 to 4, the corresponding spatial center changes from the secondary crash segment to its downstream segments. For example, the secondary crash segment is the spatial center at the time window 1. While the spatial center at time windows 2, 3, and 4 become the next nearest downstream segments. It should be noted that the spatial center is set to gradually approach but does not exceed the primary crash segment. Based on various spatial relationships between primary and secondary crashes, different patterns of spatial-temporal extraction windows are further discussed in Appendix A.

Therefore, the extracted traffic flow data by such dynamic spatial-temporal windows could reflect turbulent traffic flow dynamics among secondary crash and primary crash segments. Finally, there were 9 (variables)*4 (time windows) *3 (space windows) =108 traffic flow variables for secondary crash prediction. For instance, "Avg_Speed_1_up" means the value of average speed at upstream segment at the time window 1.

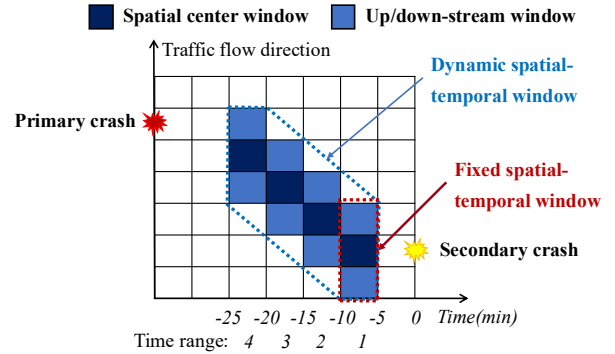

(a) Illustration of dynamic spatial-temporal window

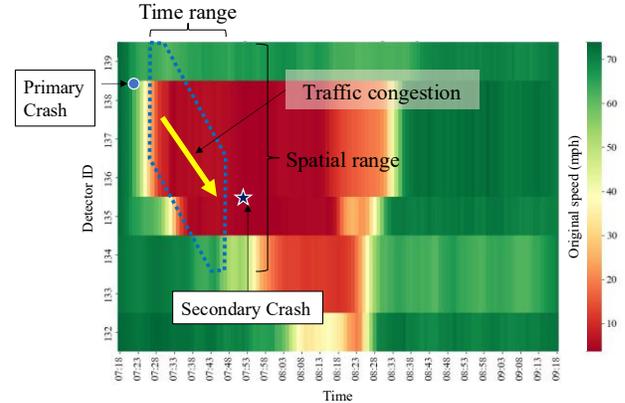

(b) Dynamic spatial-temporal window in a real case

**Fig. 5.** Dynamic spatial-temporal windows for traffic extraction.

**2) SC models under two case-control scenarios**

Based on the occurrence of secondary crash, traffic status at upstream segments actually presents three distinct cases: secondary crash case, normal crash case, and crash-free case as shown in Fig 6. Given that, the likelihood of secondary crash can be evaluated not only by comparing the traffic patterns preceding secondary crashes to those under normal crashes, but also by effectively capturing the differences between secondary crash and crash-free cases. Previous method only considered the former, resulting in biased fitting and limited prediction performance [2]. To address this issue, two distinct SC models were developed under different matched case-control rules:

**(a) SC model 1: secondary crash *vs* normal crash**

This model aims to predict the secondary crash likelihood though comparing the turbulent traffic flow with traffic statues after a normal crash. The positive samples ($y = 1$) are the secondary crash cases, while the negative samples ($y = 0$) are the normal crash cases which were created using the matched case-control design. The location of the crash and the hour of the day were used as control factors. For example, if a secondary crash happened at 7:50 on segment 2 after a primary crash on segment 1. For the same hour of the day, the non-secondary crash samples are created when a crash happens on segment 1 but does not cause a secondary crash on segment 2.

**(b) SC model 2: secondary crash *vs* non-crash**

This model focus on predicting the crash likelihoods at the secondary crash segment compared with crash-free statues. Thus, the negative samples ($y = 0$) become the non-crash cases as shown in Fig. 6. Another matched case-control rule was used to control factors of location of the crash, the time of the day and the day of the week. For example, if a secondary crash happened at 7:20 on Monday on segment 2 after a primary crash on segment 1. The corresponding non-crash samples are selected when there are no crashes on segment 2 for the same day of week (Monday) and time of the day (7:20).

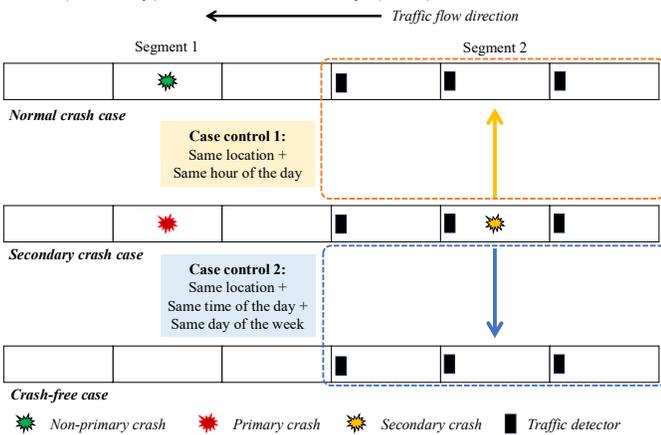

**Fig. 6.** Two case-control methods for SC models.

Based on existing studies [2, 28-31] and preliminary testing, the ratio of case-control was set as 1(positive): 4(negative) to balance rare secondary crashes. Overall, SC model 1 assesses the traffic status before secondary crashes against those at normal crashes, while SC model 2 compares the traffic status before secondary crashes with crash-free conditions.

## C. Hybrid Model

A hybrid model integrates the outputs from both the PC and SC models to generate final secondary crash likelihood. The concept behinds the hybrid model is that secondary crash likelihood could be predicted using two approaches: predicting the likelihood of crashes causing secondary crashes or predicting the likelihood of secondary crash occurrence after crashes. Moreover, the hybrid model can be seen as an ensemble learning, which combines multiple models to form a stronger predictor with better accuracy and robustness [2, 36]. For example, once the PC model fails to identify a secondary crash, two SC models can still independently provide secondary crash predictions, providing a redundant safety measure that helps reduce false positives. Similar to previous studies [2, 43, 46], the hybrid model uses a voting-based strategy: for one secondary crash case, as long as one of the three models predicts it as a secondary crash, the hybrid model will predict it as a secondary crash.

## D. ML Ensemble Model in PC and SC models

In this study, six widely utilized ML models were selected for the development of PC and SC models, including: Support Vector Machine (SVM), Random Forest (RF), eXtreme Gradient Boosting (XGBoost), Light gradient boosting machine (LGBM), the Multilayer Perceptron network (MLP), and Convolutional Neural Network (CNN). The architectures of the MLP and CNN models follow the designs in Yu et al. [36]. Finally, a max-voting ensemble of these ML models were used [43-45]:

$$\tilde{p}_i = Max(p_i^{SVM}, p_i^{RF}, p_i^{XGBoost}, p_i^{LGBM}, p_i^{MLP}, p_i^{CNN}) \quad (2)$$

$$\tilde{y}_i = \begin{cases} 1, & \tilde{p}_i > p_{thres} \\ 0, & \tilde{p}_i \le p_{thres} \end{cases} \quad (3)$$

where $\tilde{y}_i$ denotes the ensemble crash likelihood for sample $i$, $p_i^k$ means the predicted crash likelihood by the model $k$ (e.g., SVM, RF, XGBoost, etc.). If the $\tilde{p}_i$ exceeds the $p_{thres}$, the predicted label for this sample will be 1 (i.e., crash).

## V. RESULTS

### A. Experiment Design

The proposed primary crash and secondary crash prediction models were trained and evaluated according to Fig. 7. First, the prepared data was randomly divided into the training data (70%) and test data (30%). The split situation of the three models' positive/negative samples is shown in Table III. Table IV shows the final hyperparameters of six ML models. Finally, the ensemble model of the fine-tuned ML models was evaluated on the unseen test data.

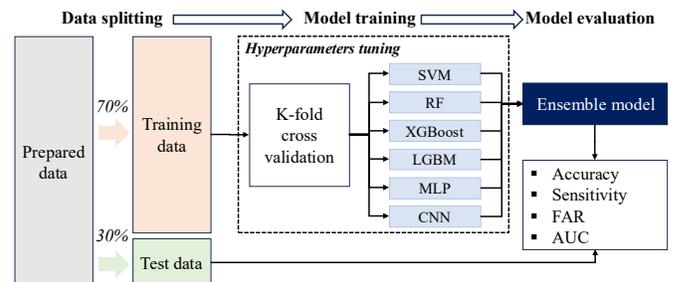

**Fig. 7.** Model development flowchart.

TABLE III
TRAINING AND TEST SAMPLES OF THREE MODELS

| Models | Positive/negative cases | Train | Test |
|---|---|---|---|
| Primary crash prediction | Primary crash | 330 | 156 |
| | Normal crash | 11931 | 5099 |
| Secondary crash prediction 1 | Secondary crash | 378 | 135 |
| | Non-secondary crash | 768 | 357 |
| Secondary crash prediction 2 | Secondary crash | 353 | 154 |
| | Non-crash | 1261 | 538 |

TABLE IV
THE OPTIMAL MODEL HYPERPARAMETERS SETTING

| | Hyperparameter | Value |
|---|---|---|
| SVM | Kernel | Radial basis kernel |
| | Penalty Parameter C | 10 |
| RF | # of estimators | 500 |
| | Max depth | 8 |
| | Min sample leaf | 4 |
| | Min sample split | 5 |
| XGBoost & LGBM | # of estimators | 400 |
| | Max depth | 4 |
| | Learning rate | $1 \times 10^{-2}$ |
| MLP & CNN | # of hidden layers in MLP | 2 |
| | # of convolution Layers in CNN | 2 |
| | convolution kernel size in CNN | (3,3) |
| | Optimizer | Adam |
| | Learning rate | $2 \times 10^{-5}$ |
| | Batch size | 32 |
| | Dropout rate | 0.2 |

*B. Models Performance Comparison*

To evaluate the model performance, $p_{thres}$ is determined at a constant false alarm rate (FAR) of 0.20 to ensure fair comparison across different models [39, 40]. Accuracy (ACC), sensitivity (SEN), and the Area Under ROC curve (AUC) were used to reflect model performance. Table V shows the prediction results of different models on the test dataset. Overall, RF model performs the best in single models, while the ensemble model slightly outperforms RF model in the three crash prediction models:

For the PC model, the ensemble model shows the best sensitivity (0.356) and AUC (0.654) while the accuracy is not improved. However, although the FAR has been controlled to be little (0.20), the overall sensitivity and AUC is relatively low. It may be due to the crash-related features such as crash severity and duration are very important variables for the occurrence of secondary crash. Thus, it would be difficult to classify one crash into primary crash or not without such detailed crash information, which also highlights the importance of secondary crash risk prediction models.

For the SC model 1, the ensemble model achieves the highest accuracy (0.744), sensitivity (0.563), and AUC (0.744). Similarly, for the SC model 2, the ensemble model also achieved the highest accuracy (0.824), sensitivity (0.844), and AUC (0.902). More than 80% of secondary crash samples could be correctly classified, which is much better than the primary crash prediction model and secondary crash model 1.

As shown in Fig. 8, the proposed hybrid model significantly improved the model performance in predicting secondary crash likelihood. the ACC was improved from 0.794/0.744/0.824 to 0.854; The sensitivity was significantly improved from 0.359/0.563/0.844 to 0.910. And the AUC was improved from 0.654/0.744/0.902 to 0.952. Compared to the best single model (SC model 2), hybrid model correctly identifies an additional 13 secondary crash cases in the test dataset. To demonstrate the benefits of the hybrid model, we compared the traffic flow features between these 13 secondary crash cases (labeled as "Hybrid only") and others correctly classified by both hybrid model and SC model 2 (labeled as "Common"). As shown in Figure 9, Hybrid only samples exhibit significantly higher average speeds and considerably lower average occupancy compared to the common samples. These findings indicate that the SC model 2 performs well in the low-speed and high-occupancy cases, which are typical congested traffic conditions after primary crashes. In contrast, for high-speed and low-occupancy scenarios, the hybrid model can leverage predictions from multiple models to correct the predictions. Therefore, the hybrid model demonstrates greater robustness to these uncommon secondary crash precursors, which is consistent with existing studies [2, 38].

TABLE V
PREDICTION PERFORMANCES OF DIFFERENT MODELS

| Models | PC model | | | SC model 1 | | | SC model 2 | | |
|---|---|---|---|---|---|---|---|---|---|
| | ACC | SEN | AUC | ACC | SEN | AUC | ACC | SEN | AUC |
| SVM | 0.778 | 0.308 | 0.565 | 0.717 | 0.430 | 0.684 | 0.785 | 0.769 | 0.789 |
| RF | **0.794** | 0.340 | 0.632 | 0.732 | 0.533 | 0.731 | 0.810 | 0.825 | 0.880 |
| XGBoost | 0.785 | 0.308 | 0.600 | 0.732 | 0.481 | 0.724 | 0.799 | 0.781 | 0.874 |
| LGBM | 0.792 | 0.321 | 0.612 | 0.703 | 0.407 | 0.722 | 0.801 | 0.788 | 0.870 |
| MLP | 0.782 | 0.288 | 0.611 | 0.693 | 0.407 | 0.689 | 0.793 | 0.775 | 0.861 |
| CNN | 0.784 | 0.302 | 0.610 | 0.720 | 0.459 | 0.709 | 0.802 | 0.794 | 0.875 |
| Ensemble Model | **0.794** | **0.359** | **0.654** | **0.744** | **0.563** | **0.744** | **0.824** | **0.844** | **0.902** |

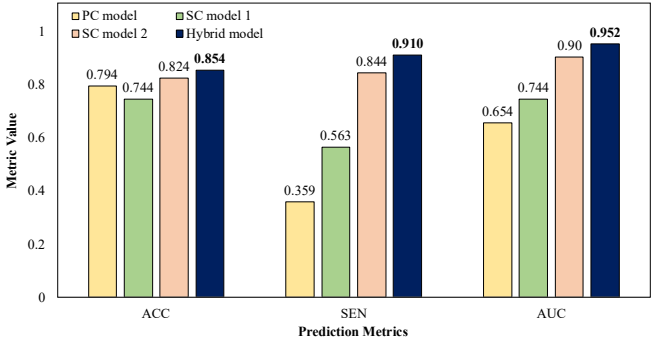

**Fig. 8.** Model performances comparison between separate and hybrid model.

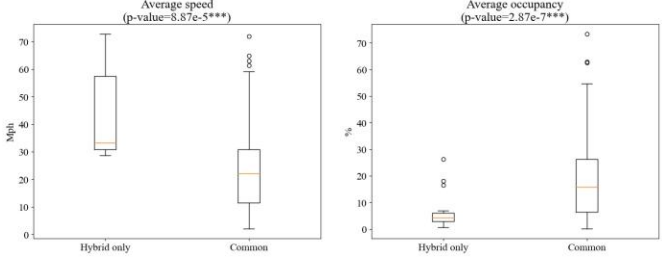

**Fig. 9.** Traffic flow feature comparison between hybrid only and common secondary crash samples.

### C. Model Interpretability Analysis

To analysis the contributing factors to the secondary crash likelihood, SHapley Additive exPlanations (SHAP) is utilized to explain the results of each ensemble model. The SHAP distributions of top-10 important variables in each crash prediction model are shown in Fig. 10 (a), (b), (c). In each subplot, contributing factors are ranked in descending order (from top to bottom) according to their importances. From the results, it can be seen that:

*Primary crash prediction model:* This model is trained for primary crashes to predict whether they have the potential to induce a secondary crash. Fig. 10(a) shows that road geometry features hold top positions in terms of their impact on primary crash likelihood. Among these, the length of the downstream segment (Miles_down) is the top-1 important feature. As its value increases (from blue to red), the associated SHAP values become positive, indicating that a crash at a location with a longer downstream segment is more likely to result in a secondary crash. One possible reason is that a longer downstream segment may accumulate more affected vehicles and take more time to clear crashed vehicles, thereby prolonging congestion and increasing the likelihood of a secondary crash. Interestingly, the variables representing the basic segment type of crash and its up- and down-stream segments (Segment_Basic_crash/down/up) exhibit positive SHAP values for primary crash likelihood. This is because, unlike other segment types where vehicles can exit (e.g., diverge or weaving areas), vehicles cannot leave freeway basic segments once a crash occurs, making congested and disordered traffic easily spread throughout the entire segments.

*Secondary crash prediction model 1:* This model aims to directly assess the secondary crash risk on upstream segments. Fig. 10(b) reveals that traffic dynamics on these segments are the most important contributors to cause secondary crashes. The most important one is the 'Cv_speed_1_center', the variation coefficient of traffic speed at the center segment 5-10 min prior secondary crashes. It represents the volatility degree of traffic speed has a significantly positive effect to the secondary crash likelihood, indicating that secondary crashes are more likely to occur when one segment has a highly fluctuating traffic speed. Overall, the result presents that the likelihoods of secondary crashes are highly related with low average speed and high coefficient of variation of speed, which exactly reflect the congested and turbulent traffic caused by the primary crashes.

*Secondary crash prediction model 2:* Similarly, traffic flow features in the spatial-temporal windows 1 and 2 serve as the top-10 significant variables. Half of them are average speeds of the center, up- and down-stream segments (i.e., Avg_speed_1_center/up/down), which show negative relationships with secondary crash likelihood. It indicates that compared to normal crash-free status, the congestion caused by the primary crash would propagate to its upstream segments and significantly increase the secondary crash likelihood. Meanwhile, results suggests that secondary crashes are more likely to occur when one segment and its upstream segments have high volatility of traffic speed and occupancy.

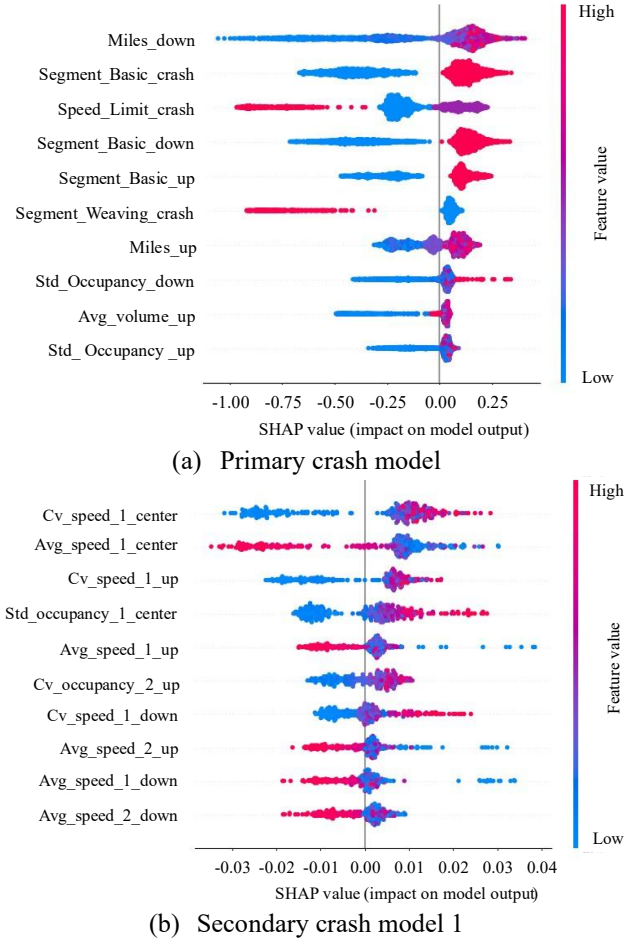

(a) Primary crash model

(b) Secondary crash model 1

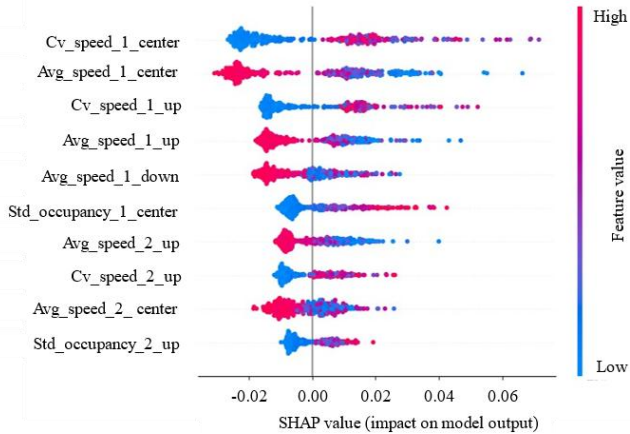

(c) Secondary crash model 2

**Fig. 10.** Top-10 important contributing factors variables.

Beyond SHAP interpretability, prediction consistency is examined to ensure trust in the ensemble model [42, 43]. Taking the SC model 1 as an example, Fig. 11 compare the predicted probabilities among individual ML models in the ensemble model. The results show that six models provide highly consistent predictions with high pairwise correlation coefficients (e.g., 0.63-0.88). For instance, the XGBoost, LGBM, and RF models yield similar prediction probabilities (approximately 0.79-0.88), and the MLP and CNN also exhibit consistent outputs with the correlation coefficients of 0.79. Similar trends are also observed in PC model and SC model 2. These findings indicate that each model has learned meaningful crash precursors and thereby generate consistent and reliable predictions, ensuring the trustworthiness of the ensemble model for real-time safety-critical system.

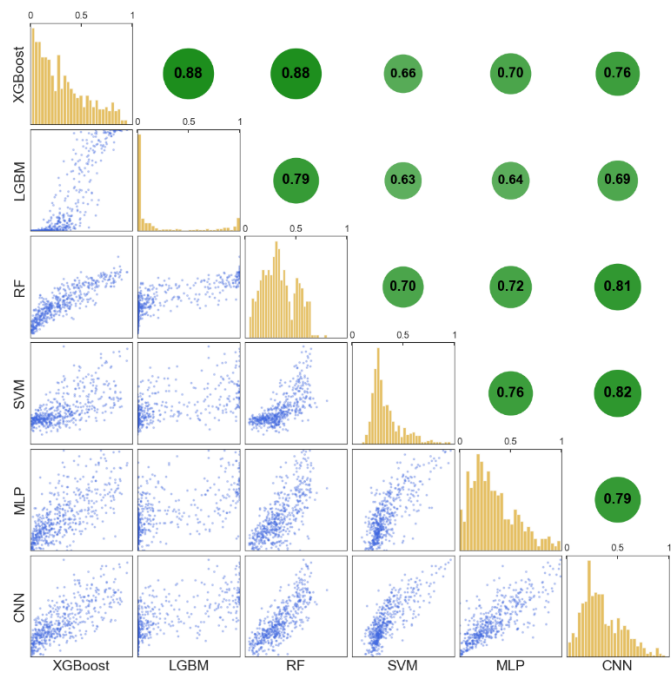

**Fig. 11.** Prediction probabilities of six models in SC model 1.

### D. Ablation Experiments

To explore the benefits of dynamic spatial-temporal windows and the optimal window size, secondary crash prediction models with different window sizes from 1 to 6 were compared in Fig. 12. Note that when the window size =1, it reduces to the traditional fixed spatial-temporal window. The results show that model performance gradually improves as the size increases and peaks at the window size of 4 (i.e., 5-25min before the secondary crash). No further improvement is observed with size of 5 and 6 (even worse). Therefore, the window size of 4 is optimal, which improves AUC by 5.13-5.66% compared with the fixed spatial-temporal window. Results indicate the benefits of the proposed dynamic spatial-temporal windows that capture more turbulent traffic flow dynamics among the secondary and primary crash segments. While in the traditional fixed spatial and temporal windows, it only focuses on the traffic status around secondary crash segments to ignore the process of backward propagation of traffic congestion from the primary crash location. However, prolong windows (e.g., 5-35min before the secondary crash) are not beneficial to improve the model performance.

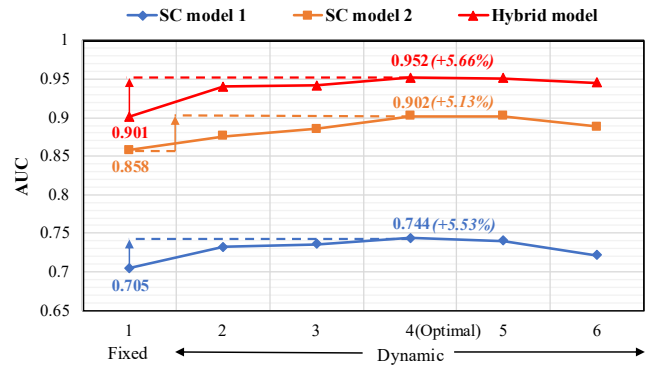

**Fig. 12.** Model AUCs of different spatiotemporal windows.

Table VI further analyzes the impacts of different types of features on the secondary crash prediction. Taking AUC value for model comparison, main findings can be summarized:

1) For the primary crash prediction, model AUC is improved from 0.601 to 0.622 and 0.642 when the weather and geometric features are included, respectively. It indicates the benefits of introducing these additional features, which is consistent with existing studies [1, 4, 29].

2) For two SC models, models with weather and geometric features also show better AUCs than only traffic flow features. Specifically, the models with weather features achieve higher AUC improvements, which indicates that real-time weather features (e.g., visibility and humidity) are more important. It may because that adverse weather conditions would lead to poor visions and slow reactions for drivers, therefore increasing the likelihood of secondary crashes [1, 47].

3) The hybrid model with only traffic flow features can provide a high AUC of 0.901. It shows that the decisive role of traffic flow status and the good benefits of ensemble of different models in secondary crash prediction. With the weather and geometric features, the model AUC are improved to 0.928 and 0.915, confirming the advantages of integrating the weather and road geometry information in hybrid models.

TABLE VI
AUC COMPARISON OF MODELS WITH DIFFERENT VARIABLES

| Model | PC model | SC model 1 | SC model 2 | Hybrid model |
|---|---|---|---|---|
| **Traffic flow (TF)** | 0.601 | 0.711 | 0.849 | **0.901** |
| **TF + Weather** | 0.622 | 0.734 | 0.883 | **0.928** |
| **TF + Geometrics** | 0.642 | 0.720 | 0.859 | **0.915** |
| **All features** | 0.654 | 0.744 | 0.902 | **0.952** |

## VI. DISCUSSIONS

### A. Different Ensemble Method Analysis

Beyond max-voting, five additional ensemble strategies are implemented to investigate their impact on model performance:

- **Mean/Median voting:** The final prediction is obtained by taking the mean or median of the predicted probabilities from all base ML models.
- **Stacking (*Heterogeneous ensemble*):** As shown in Fig. 13, stacking has two stages: Six ML models are first trained as base-learners. Their predicted probabilities are then used as input to train a meta-learner in second stage. Such stacking learns an optimal combination of individual predictions to enhance overall prediction and has been widely used in recent studies [45, 46]. Three meta-learners are used: linear regression (LR), logistic regression (Logit), and XGBoost.

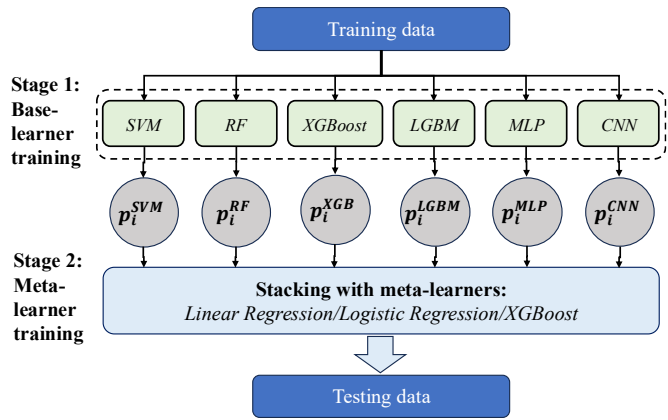

**Fig. 13.** Stacking ensemble of six ML models.

TABLE VII illustrates the results of each model under different ensemble strategies. Overall, compared with the best single model, almost all ensemble variants yield higher AUC values. However, the differences among ensemble strategies are modest. For example, in the PC model, the stacking ensemble with linear regression shows the highest AUC of 0.658, which is quite close to that of max and mean voting (0.654 and 0.657, respectively). Similar patterns are also observed in SC model 1 and 2. It is likely because the predicted probabilities from the six ML models are highly positively correlated (see Fig. 11), so simple max or mean voting is already sufficient to extract most of the ensemble gains and approach the optimal performance.

TABLE VII
MODEL AUCS UNDER DIFFERENT ENSEMBLE STRATEGIES

| Model | PC model | SC model 1 | SC model 2 |
|---|---|---|---|
| Best single model | 0.631 | 0.733 | 0.878 |
| Max voting | 0.654 | **0.744** | 0.902 |
| Mean voting | 0.657 | 0.739 | **0.905** |
| Median voting | 0.653 | 0.740 | 0.891 |
| Stacking: LR | **0.658** | 0.741 | 0.904 |
| Stacking: Logit | 0.657 | 0.742 | **0.905** |
| Stacking: XGBoost | 0.619 | 0.730 | 0.893 |

### B. Model Performance Considering Spatial-Temporal effects

According to prior studies [47, 48], crash risk may also be affected by different spatial-temporal factors (e.g., locations, and weekdays). To analyze such temporal and spatial effects, Fig. 14 visualizes the spatiotemporal distributions of secondary crashes. Secondary crashes exhibit no specific spatial patterns across different freeways. In contrast, their temporal distribution shows clear morning and afternoon peaks on weekdays but no distinct pattern on weekends.

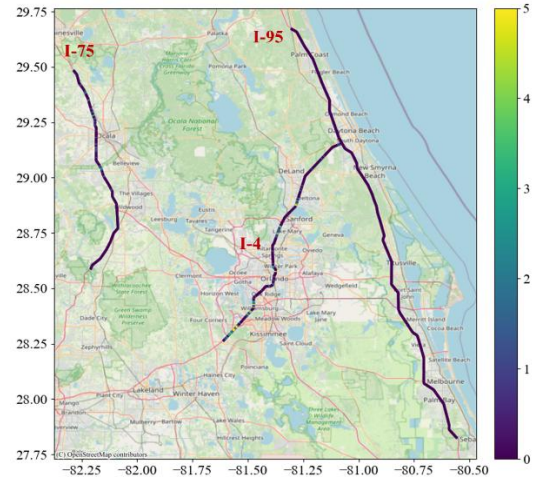

(a) Spatial distribution

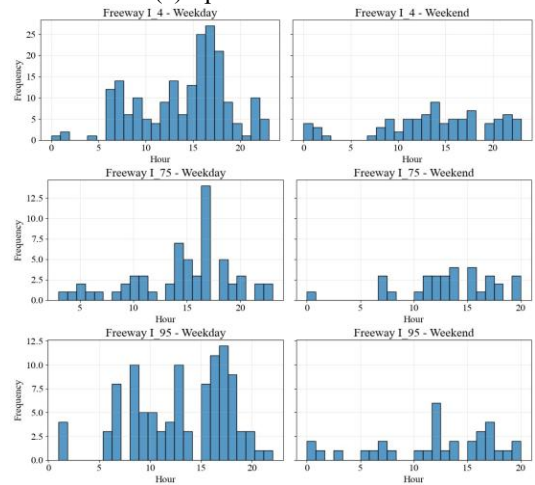

(b) Temporal distribution

**Fig. 14.** Spatial-temporal distribution of secondary crashes.

To further quantify such spatial-temporal effects on model performance, separate models for each freeway are developed to account for spatial effects, and weekday/weekend models are trained to capture temporal effects following prior studies [47, 48]. The corresponding results are reported in Table VIII and Table VIIII, respectively.

1) From Table VIII, the model performances across three freeways are generally comparable. The I-95 model has a higher AUC at SC model 2 than other freeways. However, the overall model (using all freeway samples) shows the highest performances at PC model, SC model 1, and the hybrid model. These findings suggest that spatial effects at the freeway level are unpronounced and do not substantially alter model performance, which is consistent with the lack of strong spatial patterns observed in Fig. 14.

2) From Table VIIII, for the PC model, the AUC of the weekend model is lower than that of weekday. While for the SC model 2 and the hybrid model, the AUC of weekend model is relatively higher, a pattern also reported by [47]. It indicates that traffic statues on weekdays and weekends may have difficult impacts on the occurrence of secondary crashes. Therefore, the temporal effects should be considered during the real-time secondary crash prediction and prevention.

#### TABLE VIII
#### MODEL AUC COMPARISON OF DIFFERENT ROADWAYS

| Model | Freeways | | | Overall |
|---|---|---|---|---|
| | I-4 (284*) | I-75 (93*) | I-95 (136*) | (513*) |
| PC model | 0.643 | 0.624 | 0.644 | **0.654** |
| SC model 1 | 0.732 | 0.723 | 0.720 | **0.744** |
| SC model 2 | 0.870 | 0.891 | **0.931** | 0.902 |
| Hybrid model | 0.920 | 0.921 | 0.950 | **0.952** |

*: Secondary crash sample counts in each model.

#### TABLE VIIII
#### MODEL AUC COMPARISON OF WEEKDAYS AND WEEKENDS

| Model | Weekdays (363*) | Weekends (150*) | Overall (513*) |
|---|---|---|---|
| PC model | 0.651 | 0.544 | **0.654** |
| SC model 1 | **0.761** | 0.741 | 0.744 |
| SC model 2 | 0.858 | **0.904** | 0.902 |
| Hybrid model | 0.939 | **0.967** | 0.952 |

*: Secondary crash sample counts in each model.

### C. Model Implementation Case

As one part of FDOT ATTAIN program, the developed secondary crash likelihood model has been implemented in the real-time crash prediction system for I-4, I-75, and I-95 freeways in central Florida's Regional Traffic Management Center (RTMC). In the system, once a crash occurred at one segment, the secondary crash likelihood prediction model would be triggered and then conduct the real-time prediction. Fig. 15 shows a typical example of the model implementation case. If a primary crash occurred at a segment, the model would be activated at the crash segment 0 and its five upstream segments 1-5, covering a total distance of approximately 2 km. Within the following 2h, the model would update every minute for these six segments to capture all potential high-risk status for secondary crashes. Once the predicted likelihood exceeds the predefined threshold, a secondary crash warning is issued to FDOT traffic managers, thus they can take counterpart measures to actively prevent the secondary crashes and eliminate their negative impact. During the model implementation period (November 1st to December 18th, 2023), it correctly predicted 19 of 24 secondary crashes with the threshold of 0.9. The sensitivity and FAR are 0.79 and 0.34, respectively.

Since the well-trained models are used for implementation, the computation time (including data processing and model inferencing) ranges at 20-40s with an average of 38.2s for each update. Specifically, the server in FDOT is designed to continuously collect real-time freeway traffic data and update the crash prediction results every 1min. For each update, the FDOT server processes the raw data into standardized traffic variable format and feeds them into crash prediction system. The SC prediction module then uses these inputs to calculate real-time SC likelihood for the crash segments and its upstream segments. Therefore, this short computation time ensures our proposed model well meets the real-time prediction requirements.

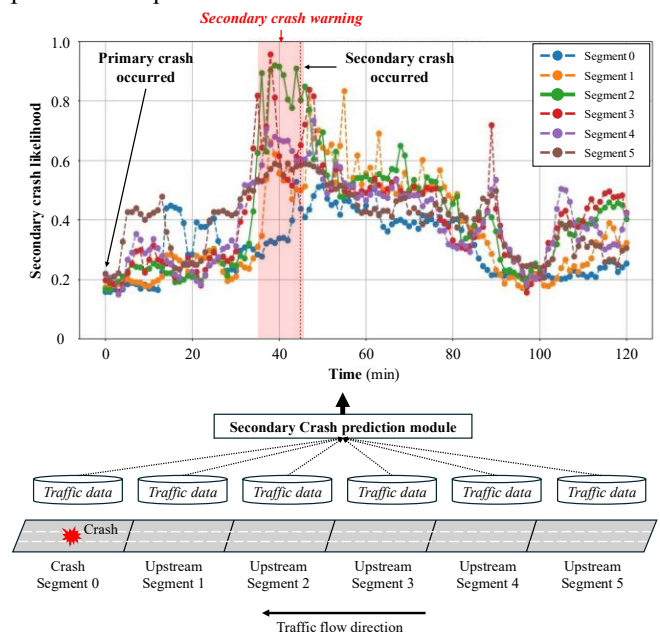

**Fig. 15.** Illustration of model implementation in real-world application.

## VII. CONCLUSIONS

Real-time secondary crash likelihood prediction has been developed to reduce response time to secondary crashes and trigger targeted proactive traffic management strategies. Most existing studies predominantly relied on the post primary crash information (e.g., crash type, severity, and duration etc.). However, these crash details are rarely available in real-time, making such methods less implementable in real world applications. Meanwhile, these studies typically focus on traffic features at the primary crash segment, neglecting those at the upstream secondary crash segments to lead poor prediction performance.

To address the above-mentioned problems, this study proposed a real-time secondary crash likelihood prediction hybrid model, which can provide high prediction accuracy without relying on any post-crash information. Four-years data from three Florida freeways were collected for the empirical experiments. A total of 513 secondary crashes, 486 primary crashes, and 20,237 normal crashes were identified for model development and test. The experimental results indicated that the proposed hybrid model could achieve outstanding performance on predicting secondary crash likelihood. It correctly identified 91% of secondary crashes with low FAR of 0.20. Compared to separate models, the AUC of secondary crash hybrid prediction was improved from 0.65/0.74/0.90 of to 0.95, which is the best result compared to the previous secondary crash prediction studies [1, 2, 18].

In conclusion, the proposed hybrid model integrates predictions of primary and secondary crashes to achieve promising accuracy, notably without relying on primary crash-related features, thus enabling this method to be freely applied to real-time implementation. The results show that primary crashes are highly related to specific road geometric features (e.g., road length and type) while secondary crashes are more likely to occur at several traffic flow conditions such as low speed and high volatility of traffic occupancy. Moreover, the model robustness to abnormal secondary crash cases and the impact of different features were also investigated. The proposed model has been successfully implemented within a FDOT real-time proactive traffic management system and achieved good predictions for secondary crashes prevention.

For further studies, novel connected vehicle data could be collected to obtain detailed and high-resolution driving information, which can help to sensitively capture the disorder in traffic conditions and improve model performance. Second, new statistical and ML algorithms can be explored to improve the prediction accuracy, consider the spatiotemporal effects [47, 49], and solve the imbalanced issue [50] between the primary crashes and normal crashes. Third, since the improvement of the ensemble model is not highly significant, exploring more advanced ML ensemble strategies may further enhance prediction accuracy and stability [51, 52]. Last but not least, due to data restriction, other freeway datasets are currently unavailable. Model generalization on other freeways would be another key direction for future research.

## APPENDIX

The spatial gap between the primary and secondary crash segments, called segment gap, differs in the samples. Four types of segment gaps were found to affect the dynamic spatial-temporal extraction windows as illustrated in the Figure A. For the segment gap is greater than or equal to 3, the spatial-temporal extraction windows are fixed as shown in Figure A. (d). The count number of the four types of cases is provided in Table A.

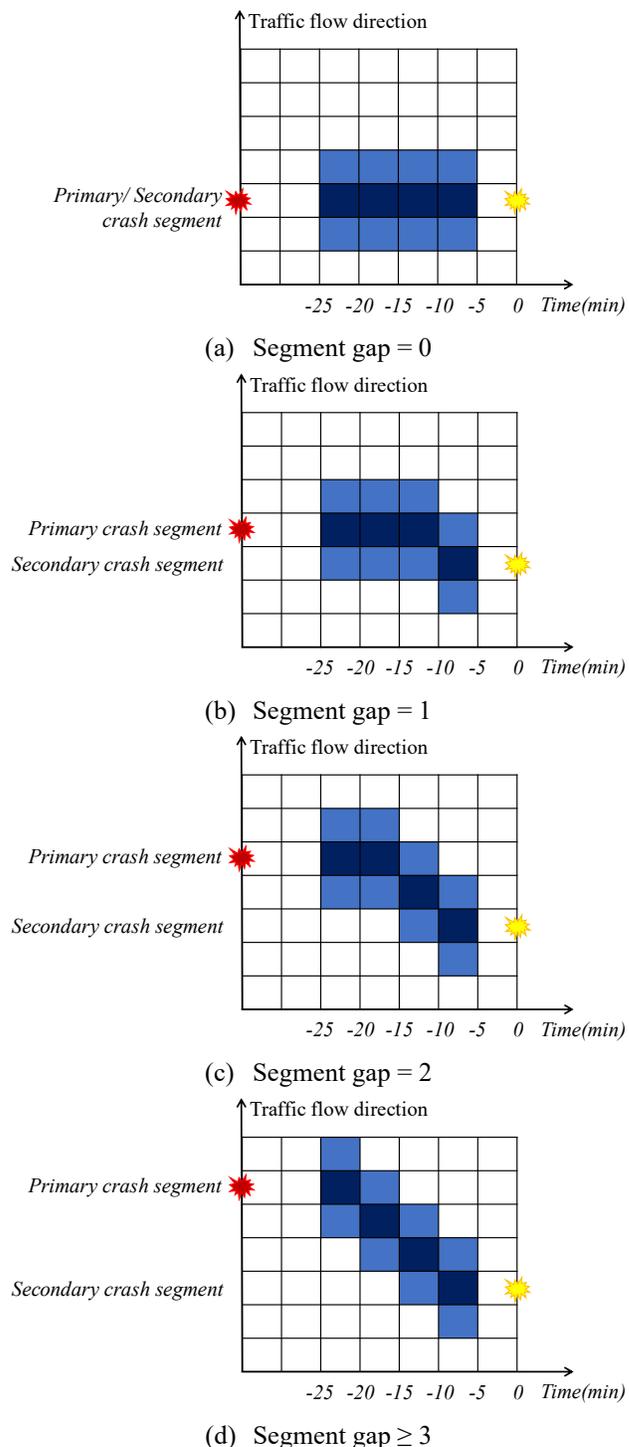

**Fig. A.** Spatial-temporal extraction windows at different segment gap.

### TABLE A
#### COUNTS OF SPATIAL-TEMPORAL EXTRACTION WINDOW CASES AT DIFFERENT SEGMENT GAPS

| Segment gap | 0 | 1 | 2 | 3-7 |
|---|---|---|---|---|
| Count | 111 | 49 | 40 | 95 |

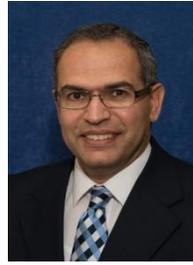

**Mohamed Abdel-Aty** (Senior member, IEEE) is a Pegasus Professor and Trustee Chair at UCF, Orlando, FL, USA. He is leading the Future City initiative at the UCF. He is also the Director of the Smart and Safe Transportation Laboratory. He has managed over 90 research projects. He has delivered more than 30 keynote speeches in conferences around the world. He has published more than 950 papers, more than 510 in journals, Google Scholar citations: 40800, H-index: 107). His main expertise and interests are in the areas of ITS, simulation, CAV, and active traffic management. He is the Editor Emeritus of Accident Analysis and Prevention. Dr. Aty has received the 2020 Roy Crum Distinguished Service Award from the Transportation Research Board, National Safety Council's Distinguished Service to Safety Award, Francis Turner award from ASCE and the Lifetime Achievement Safety Award and S.S. Steinberg Award from ARTBA in 2019 and 2022, respectively. He has also received with his team multiple international awards including the Prince Michael Road Safety Award, London 2019.

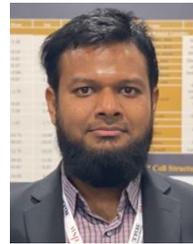

**Zubayer Islam** (member, IEEE) received the master's degree in electrical and electronics engineering from the Bangladesh University of Engineering and Technology, Dhaka, Bangladesh, in 2017, and the Ph.D. degree in transportation engineering from the University of Central Florida, Orlando, FL, USA, in 2021, where he is currently an Assistant Professor of Transportation Engineering.

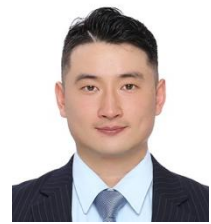

**Chenzhu Wang** received his doctor's degree in Transportation Engineering of Southeast University in 2023. He is currently working as a postdoctoral researcher in transportation engineering from the UCF. His research interests include traffic safety analysis, statistical model, intelligent transportation systems, and machine learning and deep mining.

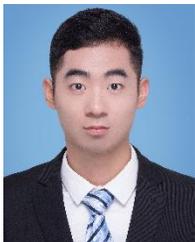

**Lei Han** received the bachelor's degree and master's degree in the College of Transportation Engineering, Tongji University in 2020 and 2023, respectively. He is currently pursuing the Ph.D. degree in transportation engineering from the University of Central Florida (UCF). He is currently a research associate with UCF. His research interests include traffic safety analysis, intelligent transportation systems, and deep learning applications in transportation engineering.